\documentclass[runningheads]{llncs}

\usepackage[T1]{fontenc}
\def\doi#1{\href{https://doi.org/\detokenize{#1}}{\url{https://doi.org/\detokenize{#1}}}}
\usepackage{graphicx}
\usepackage{hyperref}

\usepackage{listings}
\usepackage{xcolor}
\begin{document}
\title{TINC: Temporally Informed Non-Contrastive Learning for Disease Progression Modeling in Retinal OCT Volumes} 
%
%
\author{Taha Emre\and
Arunava Chakravarty\and
Antoine Rivail\and
Sophie Riedl\and\\
Ursula Schmidt-Erfurth\and
Hrvoje Bogunovi\'c}
%

\authorrunning{Emre et al.}
%
\institute{Department of Ophthalmology and Optometry, Medical University of Vienna, Austria\\
\email{\{taha.emre,hrvoje.bogunovic\}@meduniwien.ac.at}}
%
\maketitle              
\begin{abstract}
Recent contrastive learning methods achieved state-of-the-art in low label regimes. However, the training requires large batch sizes and heavy augmentations to create multiple views of an image. With non-contrastive methods, the negatives are implicitly incorporated in the loss, allowing different images and modalities as pairs. Although the meta-information (i.e., age, sex) in medical imaging is abundant, the annotations are noisy and prone to class imbalance. In this work, we exploited already existing temporal information (different visits from a patient) in a longitudinal optical coherence tomography (OCT) dataset using temporally informed non-contrastive loss (\textbf{TINC}) without increasing complexity and need for negative pairs. Moreover, our novel pair-forming scheme can avoid heavy augmentations and implicitly incorporates the temporal information in the pairs. Finally, these representations learned from the pretraining are more successful in predicting disease progression where the temporal information is crucial for the downstream task. More specifically, our model outperforms existing models in predicting the risk of conversion within a time frame from intermediate age-related macular degeneration (AMD) to the late wet-AMD stage.
\end{abstract}
\section{Introduction}
The scarcity of manually annotated labels is a major limitation for the classification tasks in medical image analysis. Self-supervised learning (SSL) showed a great promise in exploiting the availability of unlabeled medical data by outperforming models trained from random weights or pretrained with non-medical images in difficult supervised settings~\cite{azizi2021big}. Traditional SSL methods rely on pretext tasks that are believed to be semantically relevant to the downstream task, such as jigsaw-puzzle solving, and rotation angle prediction. 

In recent years, contrastive learning (CL) methods surpassed the pretext-based SSL in unsupervised representation learning. They learn similar representations of two heavily augmented views (\emph{positive pairs}) of a sample while pushing away the others in the representation space as \emph{negatives}. The goal is to find representations that are semantically meaningful and robust to image perturbations. Following this, CL methods have also been adapted for medical images. Li et al.~\cite{li2021imbalance} addressed the class imbalance problem by sample re-weighting during contrastive training and devised an augmentation scheme for 3D volumes. Chen et al.~\cite{chen2021uscl} proposed a sampling strategy by feeding two frames as pairs from an ultrasound video to encode the temporal information for the CL. They found that sampling and augmentation strategies were crucial for the downstream task. Azizi et al.~\cite{azizi2021big} showed that if two different images of a patient included the same pathology, they formed more informative positive pairs than the heavily augmented pairs. Also, they reported that if the supervised and the unsupervised data were mixed for the CL, the downstream task's performance was increased. These methods were based on contrastive InfoNCE loss~\cite{oord2019representation}. However, the success of the CL largely depends on the quality of the negative samples~\cite{ermolov2021whitening}. This introduces two challenges for medical imaging; (i) large batch sizes, and (ii) explicit negatives in a batch. Furthermore, particularly in longitudinal studies, the large batch sizes (over a thousand) are not compatible with the number of patients (a few hundred) and would create negative pairs from the same patient. 

The developments in non-contrastive learning methods~\cite{chen2021exploring,zbontar2021barlow,bardes2022vicreg} avoid the need for explicit negatives and consequently for large batch sizes. They implicitly learn to push the negatives with stop-gradient~\cite{chen2021exploring,grill2020bootstrap}, clustering~\cite{caron2020unsupervised}, or creating discrepancy between pairs through a specific loss~\cite{bardes2022vicreg,zbontar2021barlow}. Barlow Twins'~\cite{zbontar2021barlow} loss function achieves that by making the correlation matrix of the embeddings of the two views close to identity matrix and VICReg~\cite{bardes2022vicreg} by calculating a Huber variance loss within a batch of embeddings. Especially, VICReg requires no architectural trick, large batch sizes, or normalization. Both of them use a simple Siamese Network and a multi-layer perceptron (MLP) projector on the representations for the embeddings. They also allow constructing pairs from different images, and unlike contrastive methods, it is even possible to use multiple modalities.

In this paper, we focus on non-contrastive SSL in longitudinal imaging datasets, and we propose a new similarity loss to exploit the temporal meta-information without increasing the complexity of the non-contrastive training. Also, we introduce a new pair-forming strategy by using scans from different visits of a patient as inputs to the model. In this regard, our work is one of the first to build on non-contrastive learning with continuous labels (time difference between visits). 

\paragraph{Clinical background} The optical coherence tomography (OCT) imaging is widely used in clinical practice to provide a 3D volume of the retina as a series of cross-sectional slices called \textit{B-scans}. Age-related macular degeneration (AMD) is the leading cause of vision loss in the elderly population~\cite{bressAMD}. It progresses from the early/intermediate stage with few visual symptoms to a late stage with severe vision loss. Conversion to late-stage could take two forms, wet and dry AMD. Wet-AMD is defined by the formation of new vessels. An intravitreal injection can improve the patients' vision, but it is most effective when applied soon upon conversion to wet-AMD. This motivated the medical imaging research communities to develop risk estimation models for conversion to wet-AMD. AMD progression prediction in OCT has been studied using statistical methods~\cite{WU2021118,YANG20191,schmidt2018prediction} based on biomarkers and genomics. Schmidt-Erfurth et al.~\cite{schmidt2018prediction} approached the conversion prediction as a survival problem and used a cox model. Initial deep learning models works showed the importance of standardized preprocessing in B-scans~\cite{russakoff2019deep}. Recently, ~\cite{yim2020predicting} exploited surrogate tasks like retinal biomarker segmentation in OCT volumes to improve the performance. On the other hand,~\cite{Banerjee2020} used imaging biomarkers and demographics to train an RNN on sequential data. As an SSL approach, Rivail et al.~\cite{rivail2019modeling} pre-trained a Siamese network by predicting the time difference between the branches.
\paragraph{Contribution} We improve on learning representations such that they (a) capture temporal meta-information in longitudinal imaging data, (b) do not suffer from dimensional collapse (only a small part of representation space is useful~\cite{jing2022understanding}). Our contribution is two-folds. First, we proposed a simple yet effective similarity loss called TINC to implicitly embed the temporal information without increasing the non-contrastive loss complexity. We hypothesized that the B-scans acquired closer in time should be close in the representation space. We chose VICReg~\cite{bardes2022vicreg} to use TINC with because its similarity loss explicitly reduces the distance between representations and allows alternatives in its place. Second, instead of aggressive augmentation for pair generation from a single B-scan, we formed pairs by picking two moderately augmented versions of two different B-scans from a patient's OCT volumes acquired at different times. Finally, TINC had increased performance in the difficult task of predicting conversion from the intermediate to wet-AMD within a clinically-relevant time interval of 6 months.

\section{Method}
We modified the VICReg's invariance (similarity) term by constraining it with the normalized absolute time difference (no temporal order among the inputs) between the input images. The original invariance term is the mean-squared error (MSE) between two unnormalized embeddings. The time difference acts as a margin on how low the invariance term can get. As the time difference between two visits increases, the similarity between a pair of the respective B-scans should decrease. Thus, the distance measure should lie within a margin, not on a point like the VICReg invariance term does. In other terms, the representations should slightly differ due to the time difference between the scans of a patient.
\begin{figure}
    \includegraphics[width=1.0\textwidth]{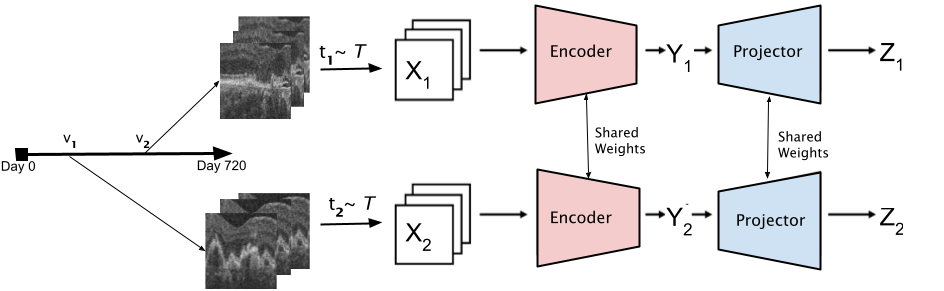}
    \caption{The overall workflow. Two B-scans sampled from different visits of a patient are fed to the network. $t_1$ and $t_2$ are the transformations for the views $X_1$ and $X_2$. An encoder produces representations $Y_1$ and $Y_2$, then a projector expands them to embeddings $Z_1$ and $Z_2$, on which the loss is calculated.}\label{vicr}
\end{figure}

Given a batch of $n$ patients with multiple visits, let visits $v_1$ and $v_2$ be the components of $n$ pair of time points randomly sampled from each available  patients' visit dates within a certain time interval. Randomly selected B-scans from the OCT volumes at times $v_1$ and $v_2$, are augmented by random augmentations $t_1$ and $t_2$ for the two views $X_1$ and $X_2$. First, the encoder produces $n$ representations $Y_1$ and $Y_2$, then the projector (also called expander) expands the representations to embeddings $Z_1$ and $Z_2$ with embedding dimension $d$ (Fig.~\ref{vicr}).

The loss terms of the original VICReg works as follows: $S$ (Eq. \ref{eq:invariance}) is the invariance term or the similarity loss, $V$ (Eq. \ref{eq:var_loss}) is the variance term to keep a variance margin between different pairs of embeddings (prevents representation collapse), and $C$ (Eq. \ref{eq:covariance}) is the covariance term that forces each component to be as informative as possible (prevents dimensional collapse).
\begin{equation} \label{eq:invariance}
    S(Z_1, Z_2) = \frac{1}{n} \sum_i \|z_{1i} - z_{2i}\|_2^2,
\end{equation}
\begin{equation} \label{eq:var_loss}
    V(Z) = \frac{1}{d} \sum_{j=1}^{d} \max(0, \gamma - \mathrm{std}(z^{j}, \epsilon)),
\end{equation}
\begin{equation} \label{eq:covariance}
   C(Z) = \frac{1}{d} \sum_{i \ne j} [Cov(Z)]_{i,j}^2, \ \ \ \textrm{where} \ \ \  Cov(Z) = \frac{1}{n - 1} \sum_{i=1}^{n} (z_{i} - \bar{z})(z_{i} - \bar{z})^{T}
\end{equation}
where in $V$, std is the standard deviation of the $z^{j}$ which is a vector of the $j$th embedding component values along the batch. For $V$, we used $\gamma$ as 1. In covariance term $C$, $Cov$ is the covariance matrix of an embedding vector. $C$ and $V$ are calculated for $Z_1$ and $Z_2$ separately. $S$ is the MSE between two vectors. Finally, the total loss is the weighted sum of these three.

\subsubsection{Temporally Informed Non-Contrastive Loss}
The temporal label is defined as the difference between visit dates $v_{1}$ and $v_{2}$. $\Delta v$ is the absolute value of the difference scaled to 0-1 using a Min-Max scaler with given $v_\textrm{min}$ and $v_\textrm{max}$. We use $\Delta v$ as a margin, where the distance between the two embeddings should not be greater than it. Inspired by the epsilon insensitive loss from support vector regression, the invariance term $S$ in VICReg is replaced with our TINC loss:
\begin{equation} \label{eq:TINC}
   \ell_{\mathrm{TINC}}(Z_1, Z_2) = \frac{1}{n} \sum_i max(0, \|z_{1i} - z_{2i}\|_2^2-\Delta v_i) 
\end{equation}
The TINC term forces the distance between the two representations to be within a non-zero margin, set proportionally to the time difference between visits. As $\Delta v$ gets close to 1, the margin becomes wider, and the loss does not enforce a strict similarity. On the other hand, when $\Delta v$ is close to 0, TINC loss becomes similar to MSE between the two representations. In principle, the values of the embedding components could diminish, resulting in a collapse. Then the distance would be smaller than the margin, not contributing to the overall loss. However, the variance term in VICReg counteracts it by enforcing a standard deviation between different pairs, preventing the components from having infinitesimal values. We kept the variance and covariance losses from VICReg unmodified.

\section{Experiments \& Results}
\paragraph{Dataset} The self-supervised and supervised trainings were performed and evaluated on a longitudinal dataset of OCT volumes of 1,096 patients \emph{fellow} eyes\footnote{The other eye that is not part of the interventional study} from a clinical trial\footnote{NCT00891735. https://clinicaltrials.gov/ct2/show/NCT00891735} studying the efficacy of wet-AMD treatment of the \emph{study} eyes. Patients had both eyes scanned monthly over two years with Cirrus OCT (Zeiss, Dublin, US) on monthly basis. The volumes consisted of 128 B-scans with a resolution of 512$\times$1024 px covering a volume of $6\times6\times2$ mm$^3$.

For the wet-AMD conversion prediction task, we selected fellow eyes that either remained in the intermediate AMD stage throughout the trial, or converted to wet-AMD during the trial, excluding those that had late AMD from baseline or converted to late dry-AMD. The final supervised dataset consisted of 463 eyes and 10,096 OCT scans with 117 converter eyes, and 346 non-converters. The rest of the eyes were included in the unsupervised dataset, which consisted of 541 eyes and 12,494 volumes. Following~\cite{yan2020deep,yim2020predicting,rivail2019modeling,Banerjee2020}, wet-AMD conversion was defined as a binary classification task, i.e., predicting whether an eye is going to convert to wet-AMD within a clinically-relevant 6 months time-frame. For the supervised training, eyes were split into 60\% for training, 20\% for validation, and 20\% for testing, stratified by the detected conversion. The evaluation is reported on the scan level.

\paragraph{Preprocessing \& Augmentations} In the supervised setting, we extracted from each OCT volume a set of 6 B-scans covering the central 0.28 mm, whereas 9 B-scans covering the central 0.42 mm were extracted for the SSL. To standardize the view, the retina in each B-scan was flattened with a quadratic fit to the RPE layer (segmentations are obtained with IOWA Reference Algorithm~\cite{10.1167/iovs.14-15669}). Then, each B-scan was cropped to $6\times0.6$ mm$^2$ and resized to $224\times224$ pixels, with intensities normalized between 0-1. All B-scans from an OCT volume were assigned the same conversion label for the supervised training. But during validation and testing, conversion probability of a volume was computed by picking the B-scan with maximum probability among the B-scans. 

\begin{figure}
    \centering
    \includegraphics[width=0.25\textwidth,height=0.25\textheight,keepaspectratio]{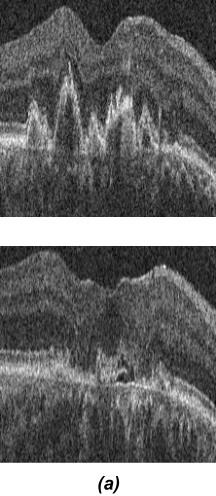}
    \includegraphics[width=0.25\textwidth,height=0.25\textheight,keepaspectratio]{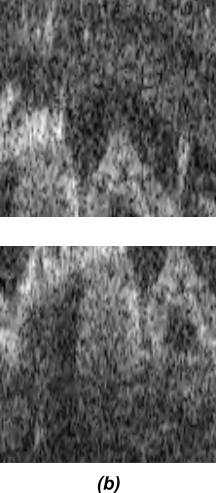}
    \includegraphics[width=0.25\textwidth,height=0.25\textheight,keepaspectratio]{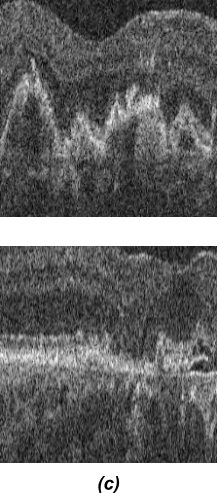}
    \includegraphics[width=0.25\textwidth,height=0.25\textheight,keepaspectratio]{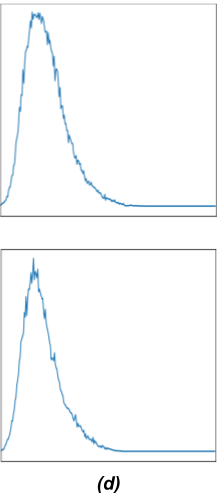}
    \caption{Two examples of different random crop strategies. \textbf{a}: flattened B-scans, \textbf{b}: small crop area ratio~\cite{bardes2022vicreg}, \textbf{c}: big crop area ratio between 0.4-0.8, \textbf{d}: color histograms of c}\label{croppyfig}
\end{figure}

For the supervised training augmentations, random translation, rotation (max 10 degrees) and  horizontal flip were used. When forming the pairs for the SSL, we followed ~\cite{bardes2022vicreg,azizi2021big} but with an increased minimum area ratio for the random cropping from 0.08 to 0.4, because in OCT volumes the noise ratio is higher than in natural images (Fig. \ref{croppyfig}), which makes small crops uninformative. Also, random grayscaling augmentation is not applicable. We picked the B-scans of the same patient from different visits with the time difference in the range of $90-540$ days. The time difference acts an additional augmentation, which makes the task non-trivial. Additionally, two large crops from two B-scans yield similar color histograms, preventing network to memorize color histograms and overfit (Fig. \ref{croppyfig}(d)). Following the protocol in~\cite{azizi2021big}, the supervised and unsupervised data were combined for the SSL.
\paragraph{Setup} ResNet-50 was chosen as the encoder backbone, and an MLP with two hidden layers with batch normalization as the projector, similar to~\cite{bardes2022vicreg} except the dimensions were chosen as 4096 for all the SSL steps. In SSL, we used AdamW with batch size of 128, learning rate of $5*10^{-4}$ and a weight decay of $10^{-6}$ for 400 epochs. Following~\cite{bardes2022vicreg,chen2021exploring}, a cosine learning rate scheduler with a warm-up of 10 epochs was used. In VICReg, the coefficient of the invariance term was fixed to 25, and we found improved performance when the coefficients of the variance and the covariance terms were set to 5 and 1, respectively. For the Barlow Twins, the coefficient for the redundancy reduction term was kept at 0.005.

For the downstream task, we provided the results from both the linear evaluation and the fine-tuning. The performance was evaluated with area under the receiver operating curve (AUROC) and the precision-recall curve (PRAUC). The linear evaluation was conducted by training a linear layer on top of the pre-trained \& frozen encoder. It is trained with Adam optimizer, batch size of 128, learning rate of  $10^{-4}$, and 5-to-1 class weights in the cross-entropy loss for 10 epochs. Fine-tuning had the same parameters with the addition of weight decay for 100 epochs. When training from scratch, the model was trained for 300 epochs. The best epoch was selected as the one with the highest AUROC score on the validation set. The learning rate was selected between $10^{-2}$ - $10^{-5}$. The weight decay was selected between $10^{0}$ - $10^{-7}$ including 0.
\subsubsection{Experiments}
We report results for wet-AMD conversion prediction task from linear evaluation and fine-tuning. We compared TINC against popular non-contrastive learning methods Barlow Twins and VICReg, which can accept different images as input pairs. When testing our new pair-forming scheme, we used the original VICReg and Barlow Twins along with their modified versions. In order to show the performance of TINC, we compared it against ResNet50 trained from scratch, VICReg and Barlow Twins modified with our new input scheme, and VICReg with additional explicit time difference prediction loss term.
\begin{table}
\caption{Linear evaluation results of SSL approaches with ResNet50 backbone. $\textrm{VICReg}_{TINC}$ is the proposed VICReg with TINC loss. "\textit{w. two visits}" indicates the model modified with new pair-forming scheme.}\label{tablin}
\centering
\begin{tabular}{|l|l|l|}
\hline
Self-supervised learning &  AUROC & PRAUC\\
\hline
VICReg~\cite{bardes2022vicreg} & Representational collapse&\\
Barlow Twins~\cite{zbontar2021barlow} & 0.686 &0.103\\
VICReg w. two visits  &  0.685 &0.085\\
Barlow Twins w. two visits &0.708&0.098\\
VICReg + Explicit Time Difference & 0.701 & 0.107 \\
\textbf{$\textrm{VICReg}_{\mathbf{TINC}}$}&  \textbf{0.738} & \textbf{0.112}\\
\hline
\end{tabular}
\end{table}

\begin{table}
\caption{Model performances for the finetuning and training from scratch}\label{tab2}
\centering
\begin{tabular}{|l|l|l|}
\hline
Method & AUROC & PRAUC\\
\hline
Backbone (random initialization) &  0.713  & 0.110\\
AMDNet~\cite{russakoff2019deep} (random initialization) & 0.676 & 0.087\\
\hline
Barlow Twins w. two visits & 0.692 & 0.091\\ 
VICReg w. two visits & 0.737 & 0.117\\
\textbf{$\textrm{VICReg}_{\mathbf{TINC}}$} &   \textbf{0.756} &  \textbf{0.142}\\
\hline
\end{tabular}
\end{table}
Although random crop \& resize are the most crucial augmentations for non-contrastive learning, small crop area may not be ideal for OCT images due to the loss in contextual information. To verify this,  we first tested Barlow Twins and VICReg with their original augmentations and input scheme as baselines followed by training them with the proposed pair-forming scheme and larger random crops. We observed (first section of Table \ref{tablin}) that with vanilla VICReg, its similarity loss quickly reached close to zero, and the representations collapsed. A significant improvement in performance was observed for both the methods with the proposed input scheme. The AUROC of Barlow Twins increased from 0.686 to 0.708 and VICReg achieved 0.685 AUROC score on linear evaluation.

On linear evaluation, TINC loss clearly outperformed both Barlow Twins and VICReg even after modifying them with our novel pair-forming scheme (\ref{tablin}). TINC achieved 0.738 AUROC, while modified VICReg and Barlow Twins achieved 0.708 and 0.685 respectively. TINC captures the temporal information better with its temporally induced margin based approach leading to these improvements.

With end-to-end fine-tuning (Table \ref{tab2}) the performance improvement due to the proposed TINC loss is more apparent, even after optimizing VICReg and Barlow Twins with our input pair scheme. We also compared our results against AMDNet~\cite{russakoff2019deep}, an architecture specifically designed for 2D B-scans. Interestingly, AMDNet could not outperform a ResNet-50 initialized with random weights.


In order to demonstrate that the temporal information is crucial in the conversion prediction task, we added time difference as an additional loss term to VICReg. For this, we concatenated the two embeddings and fed them to an MLP to obtain the time difference predictions. The labels are calculated as $v_1 - v_2$ w.r.t input order and scaled between -1 and 1. The MSE between the labels and the predictions is added besides the other VICReg loss terms. This clearly improved AUROC (Table \ref{tablin}, line 1-4), but the additional term increased the complexity of VICReg training. Whereas TINC implicitly uses the temporal information with its margin so that the representation capture the anatomical changes due to the disease progression. The experiments demonstrated that the TINC approach performs better in the downstream AMD conversion prediction task than adding the time difference as a separate term.

Additionally, we modified our loss as a squared epsilon insensitive loss to have smoother boundaries, but it degraded the performance. With Barlow Twins, we observed that the fine-tuning AUROC performance was worse than its linear evaluation by 0.016. This can be explained by the fact that in Barlow Twins, the fine-tuning reached the peak validation score within 10 epochs, same as the number of linear training epochs.
\section{Discussion \& Conclusion}
The temporal information is one of the key factors to correctly model disease progression. However, popular contrastive and non-contrastive methods are not designed specifically to capture that. Additionally, they require strong augmentations to create two views of an image, which are not always applicable to medical images. We proposed TINC as a modified similarity term of the recent non-contrastive method VICReg, without increasing its complexity. Models trained with TINC outperformed the original VICReg and Barlow Twins in the task of predicting conversion to wet-AMD. Also TINC is not task or dataset specific, it is applicable to any longitudinal imaging dataset. Moreover, we proposed a new input pair-forming scheme for OCT volumes from different time points, which replaced the heavy augmentations required in the original VICReg and Barlow Twins and improved the performance.
\subsubsection{Acknowledgements} The work has been partially funded by FWF Austrian Science Fund (FG 9-N), and a Wellcome Trust Collaborative Award (PINNACLE Ref. 210572/Z/18/Z).
%
%
%
\bibliographystyle{splncs04}
\bibliography{mybibliography}

\end{document}